\documentclass[conference]{IEEEtran}
\usepackage[T1]{fontenc}
\usepackage{xcolor}
\usepackage{parskip}
\usepackage{amsmath,amsfonts,amssymb}
\usepackage{centernot,url}
\usepackage[margin=0.75in]{geometry}
\usepackage{bm}
\usepackage{float}
\usepackage{subfigure}
\usepackage{graphicx}
\usepackage{algorithm2e}
\usepackage{algpseudocode}
\usepackage{multicol, blindtext}
\graphicspath{{./plots/}}
\usepackage{multicol}

\ifCLASSINFOpdf
\else
\fi
\begin{document}
%
\title{Big Data Regression Using Tree Based Segmentation}

\author{\IEEEauthorblockN{Rajiv Sambasivan}
\IEEEauthorblockN{Sourish Das}
\IEEEauthorblockA{Chennai Mathematical Institute\\
H1, SIPCOT IT Park, Siruseri\\
Kelambakkam 603103\\
Email: rsambasivan@cmi.ac.in, sourish@cmi.ac.in}

}


%


\maketitle

\begin{abstract}
\noindent Scaling regression to large datasets is a common problem in many application areas. We propose a two step approach to scaling regression to large datasets. Using a regression tree (CART) to segment the large dataset constitutes the first step of this approach. The second step of this approach is to develop a suitable regression model for each segment. Since segment sizes are not very large, we have the ability to apply sophisticated regression techniques if required. A nice feature of this two step approach is that it can yield models that have good explanatory power as well as good predictive performance. Ensemble methods like Gradient Boosted Trees can offer excellent predictive performance but may not provide interpretable models. In the experiments reported in this study, we found that the predictive performance of the proposed approach matched the predictive performance of Gradient Boosted Trees. 
\end{abstract}


%
\IEEEpeerreviewmaketitle

\section{Introduction and Motivation}
Regression is a common supervised learning task. As datasets get larger, scaling regression to large datasets has been an active area of research. Several ingenious methods have been developed in recent years to apply sophisticated regression techniques to datasets that are large in terms of the number of features (the big $p$ problem), or large in the number of instances (the big $N$ problem). In this work we propose a method to apply regression on datasets that are large in the number of instances. Scaling linear prediction problems (\cite{zhang2004solving}) was an important first step. However linear models may not be rich enough to provide an adequate level of performance on large complex datasets. Kernel methods(\cite{bosern}) are useful in such situations however, special techniques are needed to apply them to large datasets. Applying kernel methods to large datasets requires the solution of two problems. The first of these is to find a kernel that generalizes well. The second is to solve the computational hurdle associated with the solution. Kernel methods require inverting a matrix and this is computationally expensive ($O(N^3)$, for input of size $N$).The Nystrom method \cite{drineas2005nystrom} and the Random Kitchen Sinks \cite{rahimi2008random} are two popular methods to overcome the computational hurdle. Finding a kernel that generalizes well can be accomplished by using a kernel learning method (\cite{wilson2014thesis}) or by exploratory data analysis (\cite{duvenaud2014automatic}). Another problem with kernel methods is the choice of hyper-parameters. With small datasets, good hyper-parameter values can be learned by cross validation quickly, but with large datasets this might be impractical. Even with techniques such as the Nystrom method, experimental evaluation of each hyper-parameter value may be computationally expensive. As discussed in \cite{ghahramani_mlss_2011}, Gaussian Processes can learn the kernel hyper-parameters during training, but Gaussian Processes also have difficulties scaling to large datasets because they too require a matrix inversion ($O(N^3)$ computational cost) to compute the solution. (see \cite[Chapter 8]{rasmussen2006gaussian} and \cite{qui05}).  As is evident, scaling sophisticated techniques like kernel methods to large scale regression tasks has many challenges. A natural question to ask is if a divide and conquer strategy would alleviate any of these problems. It does and this is the approach taken in this work. A regression tree (Classification and Regression Trees (CART), \cite{breiman1984classification}) is used to segment the large dataset. The leaves of the regression tree represent the segments. Tree growth can be controlled by specifying the tree height or a minimum number of instances for the leaf nodes. In general, small leaves or large heights are associated with over fitting. The CART algorithm minimizes the variance in the leaf nodes. The leaves of the regression tree are associated with a range of response that is controlled by the height or equivalently the leaf size parameter. Depending on the size of this range and the complexity of the response within the segment, we can choose an appropriate regression technique for each leaf. A simple linear model may produce good results for some datasets while others may require a more sophisticated technique like Gaussian Process regression.\\
Ease of interpretation is an important characteristic of this approach. There are many excellent techniques that produce good predictions but do not yield an interpretable model. Ensemble methods like gradient boosted trees \cite{chen2016xgboost} or deep neural networks (\cite{hinton2006fast}) are examples of techniques that may yield good predictions but not interpretable models. In picking a modeling technique, many times we have to prioritize accuracy over ease of interpretation of the resulting model or vice versa. Of course, it is possible to have separate models for explanatory insights and predictions (\cite{shmueli2010explain}). As discussed in section \ref{sec:dor}, the approach presented in this work can produce very interpretable models. In the experiments conducted for this study, we found that estimates from the method presented in this work matched the accuracy of xgboost\cite{chen2016xgboost}. Therefore this method can produce models that are both interpretable and accurate.

\section{Problem Context}\label{sec:fda_review}
The response $\mathbf{Y}$ in regression is modeled as:
\begin{equation*}
\mathbf{\hat{Y}} = f(X) + \epsilon, 
\end{equation*} 

\noindent where $\epsilon$ is the error term. For trees, the function $f(X)$ takes the following form (\cite{friedman2001elements}):
\begin{equation}\label{eqn:tree_fn}
f(X) = \sum_{m=1}^{i=m}c_m\mathbb{I}(x \in X_m), 
\end{equation}
where:
\begin{description}
\item $\mathbb{I}(x \in X_m) = 
\begin{cases}
    1,& \text{if } x\in X_m\\
    0,              & \text{otherwise},
\end{cases}$
\item  $X_m \in \mathbb{R}^{d}$ is some region of the predictor space,
\item  $c_m$ is a constant. 
\end{description}
If we choose the squared error loss function, we want to minimize $\sum_{i=1}^{i=N}\big(\mathbf{Y_i} - \mathbf{\hat{Y}}(X_i)\big)^2$. Therefore regression trees seek a solution that minimizes the variance in the regions $X_m$. The $c_m$ in equation \ref{eqn:tree_fn} is then just the average of $\mathbf{Y_i}$ in the region $X_m$ (\cite{friedman2001elements}[chapter 9]). 

\section{Regression Trees For Segmentation}
In this work, we propose to use a regression tree for segmentation. The leaves of the regression tree represent the segments and produce a specific range of the response. Instances within a particular leaf produce similar values of the response. Tree size can be specified in terms of a height or equivalently the number of instances at the leaves (leaf size) where we stop considering further splits in tree development. The test error can be used to provide a guideline for the optimal height of the regression tree. A leaf size or height where the test error is similar to the training error indicates a tree that generalizes well. Regression trees predict a constant response for each leaf (the mean of the leaf instances). We can reject this estimate and use an appropriate regression technique to model the response in each leaf or segment. The choice of the regression technique will depend on the dataset. Noting that the number the instances in the leaf nodes is much smaller than the entire dataset, we now have the ability to use sophisticated regression techniques to model the response. We can start with simple models like linear regression and then evaluate if sophisticated techniques like kernel regression or Gaussian Process regression improve accuracy. The algorithm is summarized in Algorithm \ref{algo:reg_tree_seg_reg}.
\begin{algorithm}[ht]\label{algo:reg_tree_seg_reg}
\KwData{Dataset $\mathcal{D}$, leaf size $l_s$ and Test Dataset $\mathcal{T}$}
\KwResult{Segmented Regression Model}
\SetKwData{SM}{seg.model}
\SetKwData{SRM}{seg.reg.model}
\SetKwData{TS}{test.seg}
\SetKwData{TRM}{test.reg.model}
\SetKwData{TRP}{pred.value}
\SetKwFunction{BSM}{Fit.Seg.Model}
\SetKwFunction{BSRM}{Fit.Seg.Reg.Model}
\SetKwFunction{GSR}{Seg.Reg.Model}
\SetKwFunction{PTR}{Predict}
Remove Outliers\;
\tcc{Fit Segmentation Model}
\SM$\leftarrow$ \BSM{$\mathcal{D}, l_s$}\;
\For{Each Segment $s$ in $\mathcal{D}$}{
\tcc{Fit Segment Regression Model - can be a sophisticated model because segment size is small}
\SRM$\leftarrow$ \BSRM{$s$}\;
}
\tcc{Score Test Set}
\For{Each record $r$ in $\mathcal{T}$}{
\tcc{Get Segment for Test Record from Regression Tree}
\TS$\leftarrow$ \SM{$r$}\;
\tcc{Get regression model for \TS}
\TRM$\leftarrow$ \GSR{\TS}\;
\tcc{Obtain prediction for $r$ from \TRM}
\TRP$\leftarrow$ \PTR{\TRM, $r$}\;
}

\caption{Big Data Regression Using Regression Tree Segmentation}
\end{algorithm}

\section{Large Data Regression with Regression Trees for Segmentation}\label{sec:reg_tree_seg}
Using trees to segment data has been studied by \cite{chipman2002bayesian}. This study uses a Bayesian methodology to fit simple linear models to the leaf nodes.  \cite{gramacy2008bayesian} also applies a Bayesian approach to apply a tree model for regression on large datasets. This implementation is characterized by using priors for the trees and using Markov Chain Monte Carlo (MCMC) for posterior inference.  For small datasets, the tree structure may not generalize well. A slightly different dataset may produce a different tree. A Bayesian approach to picking the best structure in these cases is justifiable. However when datasets are large, regression trees may generalize well. This can be confirmed by examining the test set error for the regression tree. When the test set error is close to the training error, then it is reasonable to assume that the regression tree generalizes well. The tree height or the equivalently the leaf size at which we want to stop tree growth is a parameter we can control to ensure that the regression tree generalizes well. \cite{gramacy2008bayesian} uses Markov Chain Monte Carlo techniques for posterior inference. In this work we found that leaf sizes that permit us to overcome the computational hurdle of Gaussian Processes also resulted in a regression tree that generalizes well. For example with leaf sizes of around one thousand, we could compute a Gaussian Process regression model for the leaf nodes without any computational difficulty. There was no need to resort to MCMC for posterior inference.

Outliers can sometimes affect performance of regression trees (\cite{galimberti2011notes}). The effect of outliers is discussed in  section \ref{sec:eol}


\section{Effect of Outliers}\label{sec:eol}
Outliers can be problematic for some algorithms. The datasets used in this study are described in section \ref{sec:experiments}. Gradient Boosted Trees (xgboost\cite{chen2016xgboost}), Sparse Gaussian Processes \cite{titsias2009variational} and Stochastic Variational Gaussian Processes(\cite{hensman2013gaussian}) are sophisticated regression techniques for large datasets. The accuracies obtained with these techniques on the datasets used for this study without out-lier removal are shown in Table \ref{tab:without_or}.
\begin{table}[ht]
\centering
\begin{tabular}{rlrrr}
  \hline
 & Dataset & Sparse.GP & SVGP & XGBoost \\ 
  \hline
1 & Airline & 10.34 & 10.31 & 8.64 \\ 
  2 & California Housing & 0.35 & 0.34 & 0.24 \\ 
  3 & CCPP & 3.91 & 5.08 & 3.48 \\ 
   \hline
\end{tabular}
\caption{Accuracies Without Removing Outliers}\label{tab:without_or}
\end{table}
The Isolation Forest algorithm (\cite{liu2008isolation}) was then used for out-lier removal. The accuracies obtained after out-lier removal are reported in Table \ref{tab:with_or}
\begin{table}[ht]
\centering
\begin{tabular}{rlrrr}
  \hline
 & Dataset & Sparse.GP & SVGP & XGBoost \\ 
  \hline
1 & Airline & 10.33 & 9.11 & 7.94 \\ 
  2 & California Housing & 0.35 & 0.32 & 0.23 \\ 
  3 & CCPP & 3.91 & 4.61 & 3.30 \\ 
   \hline
\end{tabular}
\caption{Accuracies After Removing Outliers}\label{tab:with_or}
\end{table}
As is evident from Table \ref{tab:without_or} and Table \ref{tab:with_or}, outliers do not seem to affect the Sparse Gaussian Process algorithm, but they do affect Gradient Boosted Trees and the Stochastic Variational Gaussian Process. The best estimates were obtained using Gradient Boosted Trees.
\section{Methodology}\label{sec:methodology}
The methodology used to illustrate the effectiveness of Algorithm \ref{algo:reg_tree_seg_reg} is as follows. We apply out-lier removal using Isolation Forests (\cite{liu2008isolation}). Gradient Boosted Trees are an ensemble method that provides good performance on large scale regression tasks. This is evident from the results presented in the section \ref{sec:eol}, xgboost clearly does better than the other methods for large scale regression tasks. Therefore we use xgboost on the datasets used for this study and obtain a performance measurement (test set error). It should be noted that xgboost does not yield an interpretable model, but it may provide good estimates. In contrast, the models produced by the method presented in this work are interpretable (see section \ref{sec:interpretation}). We then use a regression tree to segment the dataset. We use a leaf size such that the regression tree is generalizable. We then  start with a simple model for the leaf nodes, like linear regression and obtain the test set error. We can experiment different leaf sizes to determine a leaf size that yields the lowest overall test error. In all experiments, we noted that the regression tree generalized well at the leaf sizes where the overall lowest test error was obtained. We can then check to see if sophisticated leaf regression models like Gaussian Process regression, improve performance.
\section{Experiments} \label{sec:experiments}
\subsection{Datasets}
The following datasets were used for this study:
\begin{enumerate}
\item \textbf{Combined Cycle Power Plant}: This dataset was obtained from the UCI Machine Learning repository (\cite{Lichman}). This dataset has 9568 instances. The target variable is the net hourly electrical power output from a power plant. The dataset has four features.

\item \textbf{California Housing}: This dataset was obtained from the LIAD repository (\cite{LIACC}). The target variable for this dataset is the median house price. The dataset has 8 features and 20460 instances

\item \textbf{Airline Delay}:This dataset was obtained from the US Department of Transportation's website (\cite{RITA_Delay_Data_Download}). The data represents arrival delays for US domestic flights during January and February of 2016. This dataset had 12 features and over two hundred and fifty thousand instances. Departure delay is included as one of the predictors while \cite{hensman2013gaussian} does not include it. Also the raw data includes a significant amount of early arrivals (negative delays). For all regression methods considered in this study, better models were obtained by limiting the data to the delayed flights only (arrival delays were greater than zero). This suggests that arrival delays and early arrivals are better modeled separately.

\end{enumerate}

\subsection{Experimental Evaluation of Leaf Size}
As discussed in section\ref{sec:reg_tree_seg} and section \ref{sec:methodology}, the leaf size (or equivalently the tree height) is an important parameter for the algorithm presented in this work. The leaf size can affect:
\begin{enumerate}
\item The generalization of the regression tree used to segment the data.
\item The generalization of the overall model.
\end{enumerate}
Therefore we need two experiments. The first experiment illustrates the effect of the leaf size on the generalization of the regression tree model. The second experiment illustrates the effect of the leaf size on the overall model (Algorithm \ref{algo:reg_tree_seg_reg}). For both these experiments, 70 percent of the data was used for training and 30 percent of the data was used for the test set. For both these experiments, the Root Mean Square Error (RMSE) was used as the metric for error.

\subsection{Software Tools}
All modeling for this study was done in \texttt{Python}. Preprocessing for the airline dataset was done with \texttt{R}. The \texttt{scikit-learn}(\cite{scikit-learn}) and \texttt{GPy} (\cite{gpy2014}) packages were used for model development.

\section{Discussion of Results}\label{sec:dor}
\subsection{Effect of Leaf Size on Regression Tree Generalization}\label{sec:eff_lrrt}
\begin{figure}[ht]
\includegraphics[width=\linewidth]{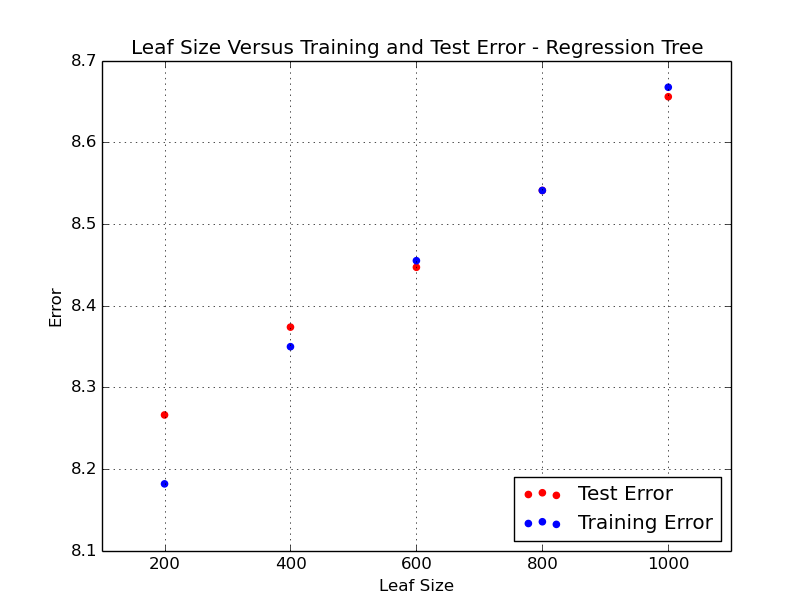}\par\caption{Airline Delay Regression Tree Generalization}
\label{fig:EXP_LSRT_AD}
\end{figure}

\begin{figure}[ht]
\includegraphics[width= \linewidth]{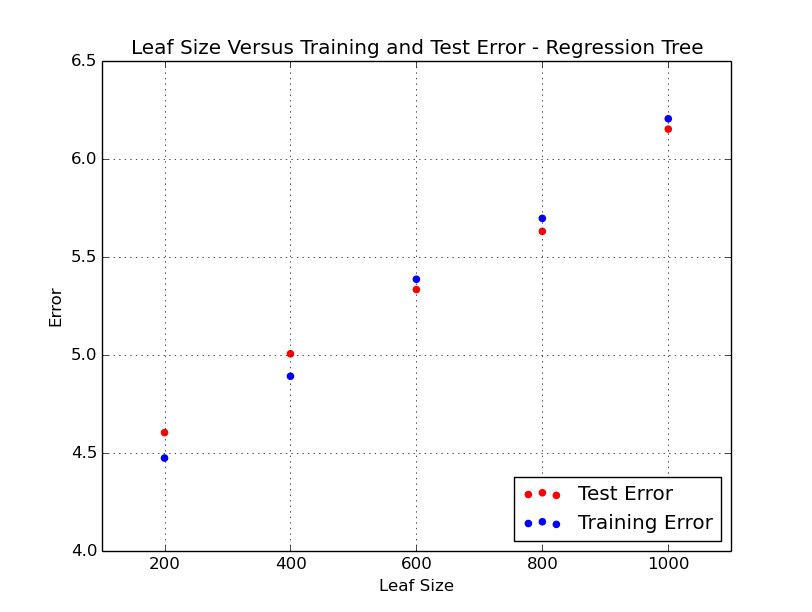}\par\caption{CCPP Regression Tree Generalization}
\label{fig:EXP_LSRT_CCPP}
\end{figure}

\begin{figure}
\includegraphics[width= \linewidth]{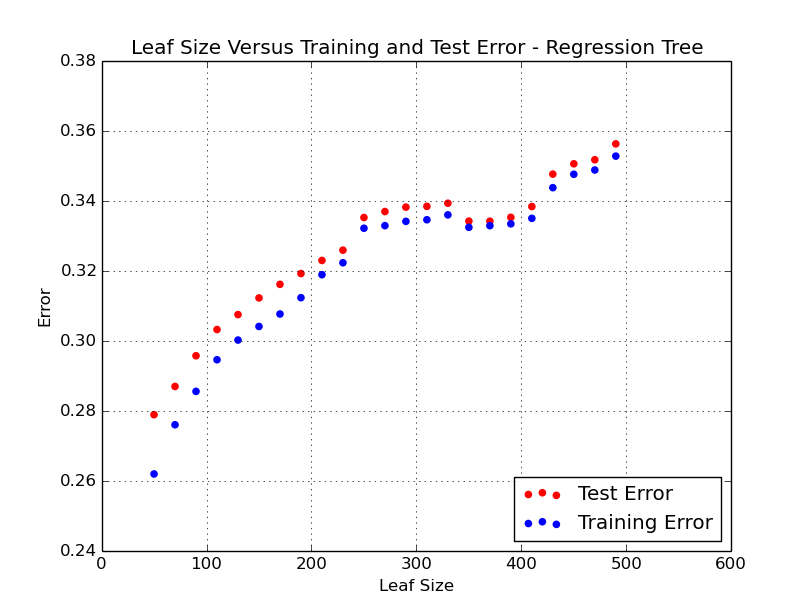}\par\caption{California Housing Regression Tree Generalization}
\label{fig:EXP_LSRT_CH}
\end{figure}

To illustrate the effect of leaf size on the generalization error of the regression tree, the tree was grown such that tree growth is stopped along a path in the regression tree when leaf size falls below the threshold value. We then measure the training error and the test error. This procedure is repeated for various values of the leaf size threshold value. The results are shown in Figure \ref{fig:EXP_LSRT_AD} through Figure \ref{fig:EXP_LSRT_CH}.

As is evident, at smaller leaf sizes (greater height of the tree), there is a greater difference between the training error and the test error. This is consistent with the idea of pruning the regression tree to prevent over fitting. Beyond a particular leaf size, increasing the leaf size does not reduce the gap between the training error and the test error.
\subsection{Effect of Leaf Size on Model Generalization}
\begin{figure}[ht]
\includegraphics[width=\linewidth]{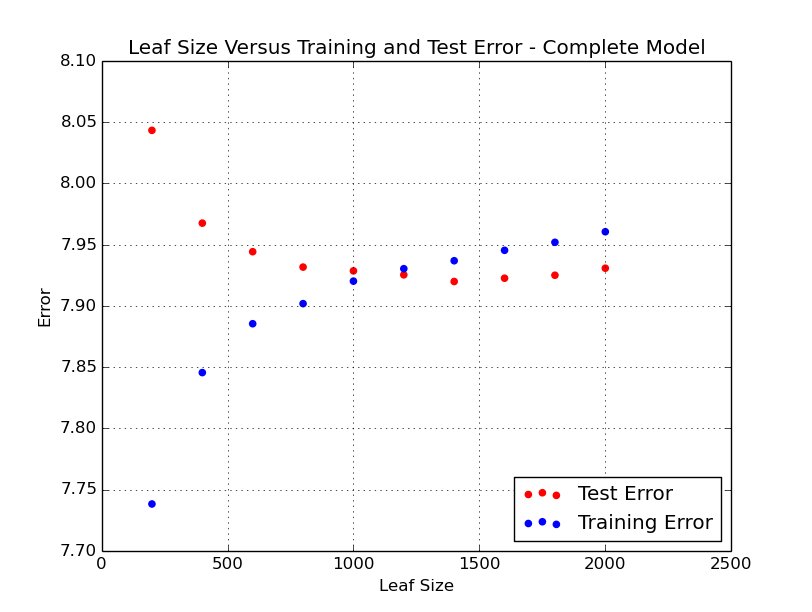}\par\caption{Airline Delay Model Generalization}
\label{EXP_LSOM_AD}
\end{figure}

\begin{figure}[ht]
\includegraphics[width= \linewidth]{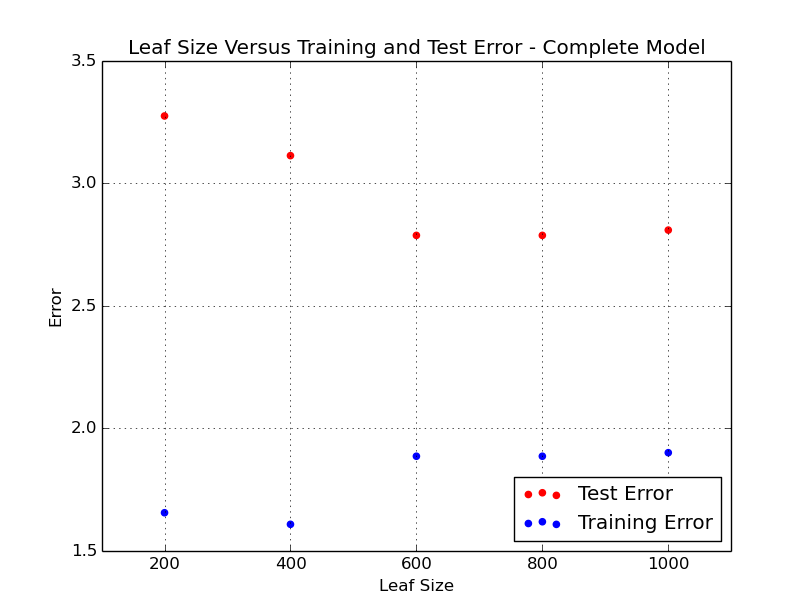}\par\caption{CCPP Model Generalization}
\label{EXP_LSOM_CCPP}
\end{figure}

\begin{figure}[ht]
\includegraphics[width= \linewidth]{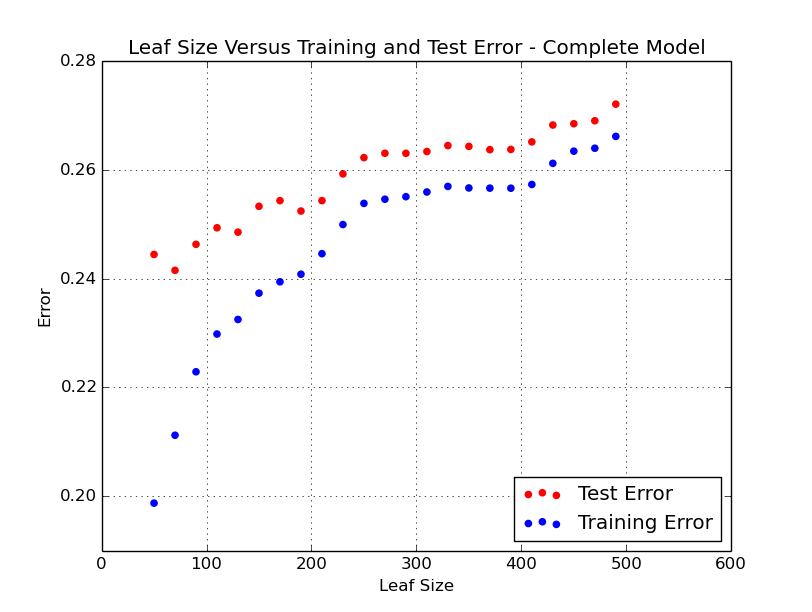}\par\caption{California Housing Model Generalization}
\label{EXP_LSOM_CH}
\end{figure}

These experiments illustrate the effect of leaf size (equivalently, the tree height) on the overall model generalization error. The overall model is one that implements Algorithm \ref{algo:reg_tree_seg_reg}. This consists of the regression tree for segmentation and an appropriate leaf regression model. For each leaf size in a set of leaf sizes, we measure the training and test errors. The results of these experiments for the datasets used in this study are shown in Figure \ref{EXP_LSOM_AD} through Figure \ref{EXP_LSOM_CH}. As with the experiments described in section \ref{sec:eff_lrrt}, smaller leaf sizes provide lesser generalizable models - with smaller leaf sizes there is a greater difference between the training error and the test error.

\subsection{Accuracy}\label{sec:accuracy}
The accuracies obtained with Algorithm \ref{algo:reg_tree_seg_reg} on the test set for the datasets used in this study are shown in Table \ref{tab:accuracy}
\begin{table}[H]
\centering
\scriptsize
\begin{tabular}{llrrr}
  \hline
 	Dataset & Leaf.Method & LS & TSR RMSE & XGBoost \\ 
  \hline
	CCPP & Gaussian Process & 1000 & 2.81 & 3.3\\ 
  	California Housing & Linear Regression &  70 & 0.24 & 0.23 \\ 
    Airline Delay & Gaussian Process & 1000 & 8.00 & 7.94\\ 
   \hline
\end{tabular}
\caption{Test Set RMSE Using Regression Tree to Segment and Leaf Regression Estimators }\label{tab:accuracy}
\end{table}
\normalsize
In Table \ref{tab:accuracy}, the Leaf Method column refers to the regression method for the segments, TSR RMSE refers to the Root Mean Square Error (RMSE) from the proposed method and the XGBoost column refers to the RMSE from  XGBoost.
The leaf size used in Table \ref{tab:accuracy} are the values at which best test set RMSE was observed. This may not be the leaf size at which the regression tree generalizes best as is evident with California Housing dataset. The best test set RMSE was obtained with a leaf size of $70$. However the regression tree model is slightly more generalizable at higher leaf sizes (see Figure \ref{fig:EXP_LSRT_CH}). Algorithm \ref{algo:reg_tree_seg_reg} permits us to pick the most appropriate leaf regression algorithm for a dataset. For this work, we started with a simple model for the leaf regression - linear regression. The test set accuracy obtained with linear regression for the leaves is noted. We then tried more sophisticated techniques to see if there was an improvement in test set RMSE. For the airline delay and California housing datasets, there was no marked decrease in test set RMSE when sophisticated regression techniques like Gaussian Process regression were used. However for the Combined Cycle Power Plant dataset, there was a marked increase in accuracy when we used a Gaussian Process with a composite kernel - sum of linear and Radial Basis Function, as the leaf regression model. 

\subsection{Interpreting the Model}\label{sec:interpretation}
\begin{figure}[ht]
\includegraphics[width=\linewidth]{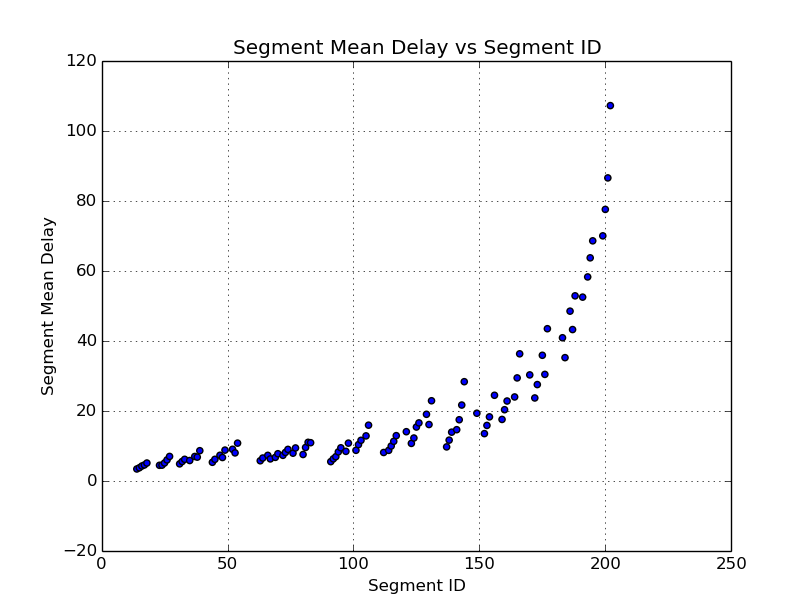}\par\caption{Segment Mean Arrival Delay}
\label{fig:SEG_MEAN_CD}
\end{figure}

\begin{figure}[ht]
\includegraphics[width= \linewidth]{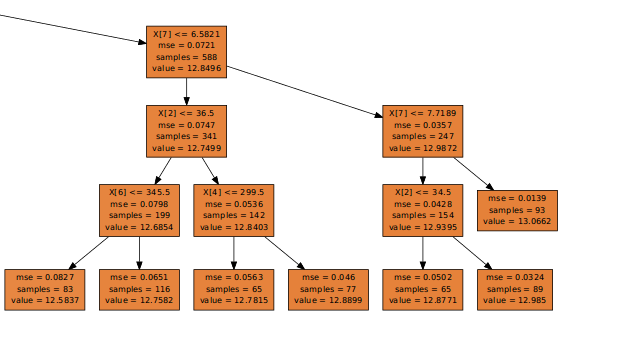}\par\caption{California Housing Tree Excerpt}
\label{fig:SEG_PROFILE}
\end{figure}

The segmentation from the regression tree model can yield some very useful insights. For example Figure \ref{fig:SEG_MEAN_CD} shows the mean segment delay for various segments for the airline delay dataset. If we know the segment a particular flight belongs to, the segment characteristics can provide useful information. For example the mean delay for that segment tells us if this flight is associated with high delays. Similarly the segments associated with the California housing dataset can provide information such as the home values associated with that segment, the median income associated with that segment etc. The leaf regression model can reveal the next level of detail associated with a particular record. For example since the leaf regression model for the California housing dataset is simple linear regression, the leaf regression model could tell us how a unit change in median income affects home values for that segment. Segment profiling can be done quite easily using the regression tree. Following the path to a segment in the regression tree, provides the segment profile for that segment. Most regression tree implementations provide a feature to visualize the regression tree. An excerpt from the regression tree for the California housing dataset is shown in Figure \ref{fig:SEG_PROFILE}

\begin{figure}[ht]
\includegraphics[width= \linewidth]{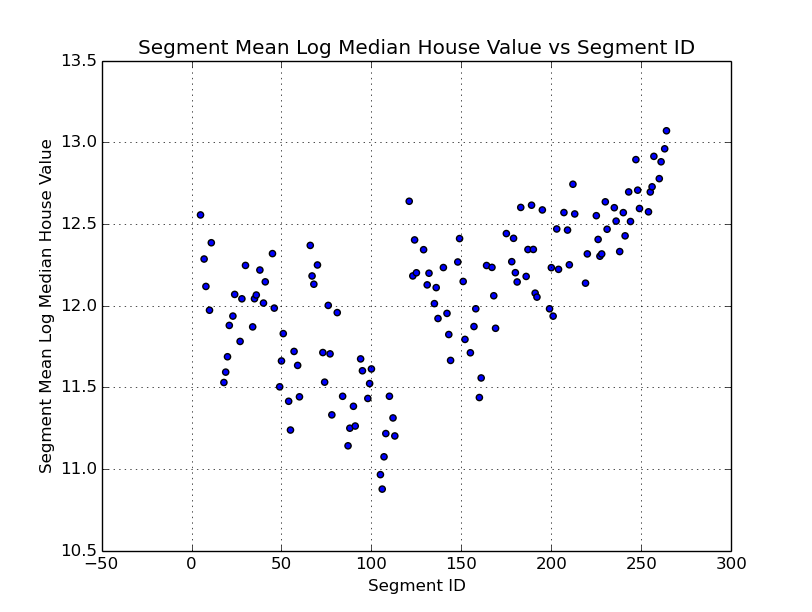}\par\caption{Segment Mean Log Median Home Value}
\label{fig:SEG_MEAN_LMHV}
\end{figure}

\section{Conclusion}\label{sec:conclusion}
In this work we presented a two step approach to large scale regression tasks. The proposed approach can produce models that have good explanatory power as well as good predictive performance. This is highly desirable. The leaf size or the tree height used for the regression tree is an important algorithm parameter. We provided experiments that illustrate the effect of this parameter on the generalization capability of the regression tree as well as the generalization capability of the overall model. Since this technique works well for regression, it is possible that this approach could be used for other large scale supervised learning tasks. This is an area of future work.




%
\medskip

\bibliographystyle{IEEEtran}

\bibliography{lsuart}

\end{document}